\documentclass{article}

\usepackage{PRIMEarxiv}

\usepackage[utf8]{inputenc} 
\usepackage[T1]{fontenc}    
\usepackage{hyperref}       
\usepackage{url}            
\usepackage{booktabs}       
\usepackage{amsfonts}       
\usepackage{nicefrac}       
\usepackage{microtype}      
\usepackage{lipsum}
\usepackage{fancyhdr}       
\usepackage{graphicx}       
\graphicspath{{media/}}     

\title{$SCORE$: A 1D Reparameterization Technique to Break Bayesian Optimization's Curse of Dimensionality} 

\author{
  Joseph Chakar \\
  Laboratoire de Physique des Interfaces et des Couches Minces (LPICM)\\
  Ecole Polytechnique, Institut Polytechnique de Paris (IP Paris)\\
  Institut Photovoltaïque d'Île-de-France (IPVF)\\
  Palaiseau, France\\
  \texttt{josephchakar8@gmail.com}\\
}

\begin{document}
\maketitle

\begin{abstract}
Bayesian optimization (BO) has emerged as a powerful tool for navigating complex search spaces, showcasing practical applications in the fields of science and engineering. However, since it typically relies on a surrogate model to approximate the objective function, BO grapples with heightened computational costs that tend to escalate as the number of parameters and experiments grows. Several methods such as parallelization, surrogate model approximations, and memory pruning have been proposed to cut down computing time, but they all fall short of resolving the core issue behind BO’s curse of dimensionality. In this paper, a 1D reparametrization trick is proposed to break this curse and sustain linear time complexity for BO in high-dimensional landscapes. This fast and scalable approach named $SCORE$ can successfully find the global minimum of needle-in-a-haystack optimization functions and fit real-world data without the high-performance computing resources typically required by state-of-the-art techniques.
\end{abstract}


\section{Introduction}
Optimization problems are ubiquitous in various fields, ranging from computer science and engineering to finance and healthcare. Whether the focus is on minimizing costs or improving efficiency, these challenges frequently involve finding the best outcome from a large pool of feasible solutions within defined constraints. Thanks to its ability to efficiently navigate this search space, Bayesian Optimization (BO) has emerged as a go-to solution to tackle these problems \cite{NIPS2012_05311655,shahriariTakingHumanOut2016,uenoCOMBOEfficientBayesian2016,kleinFastBayesianOptimization2017, alaaAutoPrognosisAutomatedClinical2018, wuHyperparameterOptimizationMachine2019, cakmakBayesianOptimizationRisk2020, liangBenchmarkingPerformanceBayesian2021, loeyBayesianbasedOptimizedDeep2022a, tripathiModelingBitcoinPrices2023}, especially in cases where running the objective function is expensive or time-consuming.

In the typical BO setup \cite{frazierTutorialBayesianOptimization2018}, a surrogate model – often a Gaussian Process (GP) regression – is leveraged to estimate the target response function from given input data. An acquisition function is then used to suggest strategic new test points based on the uncertainty level of this model. If selected carefully, this sampling strategy allows BO to explore the parameter space to uncover promising regions with high uncertainty or exploit known favorable regions to refine the search toward the global optimum. 

However, as the dimensionality of the objective function increases and more data points $N$ are needed, the time required to perform GP regression scales with $O(N^3)$ complexity \cite{snelsonSparseGaussianProcesses2005, buiUnifyingFrameworkGaussian2017, siemennFastBayesianOptimization2023} and becomes computationally intractable. In recent literature, several solutions have been proposed to lower BO’s computing time and circumvent this curse of dimensionality. A first set of solutions reduces wall-clock time by executing parallel BO evaluations \cite{munawarTheoreticalEmpiricalAnalysis2009, talnikarParallelOptimizationLarge2014, rebolledoParallelizedBayesianOptimization2020, wangParallelBayesianGlobal2020, binoisPortfolioApproachMassively2023}, but requires specialized hardware and/or software infrastructure to effectively distribute and coordinate processes across multiple computing units. Another seeks to replace the standard GP surrogate with sparse GP regression \cite{snelsonSparseGaussianProcesses2005, buiUnifyingFrameworkGaussian2017, titsiasVariationalLearningInducing2009, heEfficientBayesianOptimization2020, mossInducingPointAllocation2023} or other algorithms such as neural networks \cite{snoekScalableBayesianOptimization, NIPS2016_a96d3afe, liMultiFidelityBayesianOptimization2020, limExtrapolativeBayesianOptimization2021, li2023study}. The main drawback of changing the surrogate model is the potential loss of reliable uncertainty estimation offered by GP regression, which is crucial for decision-making and exploration/exploitation trade-offs in BO. A third set of solutions bounds the search space using trust regions that sample areas more likely to contain the optimum \cite{regisTrustRegionsKrigingbased2016, erikssonScalableGlobalOptimization2019, diouaneTREGOTrustregionFramework2023}. However, in complex optimization landscapes, these regions may not accurately capture the true shape of the objective function and thus prevent the algorithm from exploring promising regions of the search space. Additionally, determining the appropriate size and shape of the trust region is a challenging task that can require costly iterations or heuristics, which further complicates the optimization process. Similarly, memory pruning \cite{siemennFastBayesianOptimization2023} can help reduce memory usage, but it may discard useful information prematurely and hinder BO’s ability to adapt to changing conditions and exploit past experiences effectively, while still requiring considerable computational overhead.

In this work, fast and scalable BO is achieved using a 1D reparametrization technique named $SCORE$. Instead of applying the surrogate model to the entire D-dimensional space, GP regression is performed on individual 1D spaces to keep the computing time in check and speed up convergence. This is akin to projecting the D-dimensional space onto every parameter, which allows for a more manageable and efficient optimization process. Besides preventing the computational load from escalating with each iteration, applying GP regression on the parameter scale offers several advantages. It first gives the possibility to evaluate the surrogate model over full parameter ranges. In contrast, evaluating the model over the entirety of the search space is not feasible in high-dimensional spaces. This leaves standard BO approaches with no choice but to iteratively evaluate it on randomly selected points when suggesting the next set of parameters for objective function testing, potentially leading to inadequate exploration of the search space. Moreover, with $SCORE$, the approach presented herein, one can choose the number of parameter values (and consequently parameter combinations) to evaluate at every iteration. The same number of objective function evaluations can thus be split across multiple iterations with an (un)even number of runs per iteration (rather than just one), which can simultaneously reduce the number of GP regression calls and allow for faster convergence towards the optimal solution. Last but not least, all of the techniques proposed in the literature can be seamlessly integrated into $SCORE$ to further improve its performance. 

Herein, the ability of $SCORE$ to efficiently find the global minima of the 10- and 200-D Ackley functions and fit real-world current-voltage data of a solar panel is demonstrated.

\section{Bayesian Optimization}
Bayesian optimization is a well-established strategy for finding the global minimum (or maximum) of black-box functions that are expensive to evaluate.
\begin{equation}
  \begin{array}{l}
    \min f(x) \\ 
  \end{array}
\end{equation}	
where the input $x \in \Re^D$ for values of $D \leq 20$ for most successful applications \cite{frazierTutorialBayesianOptimization2018, snoekScalableBayesianOptimization}. It relies on the construction of a probabilistic (GP regression) model that serves as a computationally efficient approximation of the objective function $f(x)$ to predict the performance of unexplored points based on observed evaluations. Conditioned on a prior (Gaussian) distribution and a set of $N$ observations, this surrogate model computes a posterior distribution characterized by its mean and standard deviation. An acquisition function applied to the posterior mean and variance is then queried to balance between exploration and exploitation and reason about the next input to evaluate in the search for the global optimum of the function of interest. 

Common BO acquisition functions include the lower confidence bound (LCB), probability of improvement (PI), expected improvement (EI), and Entropy Search (ES) \cite{ NIPS2012_05311655, bullConvergenceRatesEfficient2011, hennigEntropySearchInformationEfficient2011,gelbartBayesianOptimizationUnknown2014}. These static techniques only adjust sampling based on the output of the surrogate model, but active learning-based tuning of their hyperparameters can be implemented to improve the sampling quality and prevent entrapment in local optima \cite{siemennFastBayesianOptimization2023, wangNewAcquisitionFunction2017, noeNewImprovementBasedAcquisition2018}.

While the efficacy of BO is contingent upon the choice of acquisition function, this study focuses mainly on the EI function. In minimization problems, EI is defined by:
\begin{equation}
EI(x) = (f(x^*)- \mu - \zeta)\psi(\frac{f(x^*)- \mu - \zeta}{\sigma(x)}) + \sigma(x)\phi(\frac{f(x^*)- \mu - \zeta}{\sigma(x)})
\end{equation}	
where $\psi(z)$ and $\phi(z)$ denote the cumulative distribution and density functions of a standard Gaussian distribution, respectively. $\mu(x)$ and $\sigma(x)$ are the posterior mean and standard deviation at $x$, $f(x^*)$ is the best-observed value of the objective function so far, and $\zeta$ is a free parameter to control the degree of exploration.

The rationale behind using a surrogate model in BO lies in its ability to efficiently approximate the behavior of the true function across the entire parameter space without necessitating expensive objective function evaluations. However, this surrogate model itself can become computationally intractable once the number of observations increases, especially in high dimensional spaces, which is why a proxy optimization over the acquisition function is necessary when predicting the next input to evaluate. This can lead to longer optimization times and resource-intensive computations, undermining the efficiency gains initially sought through surrogate modeling.

\section{Methods}
Inspired by the concept of separable convolutions \cite{cholletXceptionDeepLearning2017, chakarDepthwiseSeparableConvolutions2020}, $SCORE$ lifts this curse of dimensionality by decomposing the full $D$-dimensional space into $D$ one-dimensional spaces along each input variable. 

As is the case with standard BO, the objective function is first evaluated at $N_i$ initial points. Then, much like deriving the marginal probability distribution of a single variable from the joint probability distribution describing the relationship between multiple variables, each parameter is considered alone while marginalizing out the others. But instead of integrating over all the possible values of the other parameters, only the minimum (or maximum) value achieved so far for the objective function is recorded. This enables fitting the surrogate model to individual (discrete or continuous) variables and significantly reduces the computational load, which becomes dependent on the number of input variables and their mesh resolution – rather than the number of experiments $N$. In this work, only discrete variables are considered.

Next, the acquisition function is computed to assign a “score” to every possible parameter value. These individual parameter scores are then aggregated to identify the most promising parameter combination for objective function testing. Instead of suggesting just one point (i.e. the parameter combination with the best score), multiple combinations can alternatively be selected at every iteration. Therefore, for the same total number of function evaluations, the number of times the surrogate model is called is greatly reduced while potentially accelerating convergence toward the global optimum. 

\section{Results}
In this section, the computing time and efficiency of $SCORE$ using EI are benchmarked against BO \cite{vehtariGPyOptBayesianOptimization2016} on two types of optimization problems: the Ackley function \cite{siemennFastBayesianOptimization2023, erikssonScalableGlobalOptimization2019, tanTwostageKernelBayesian2023}, a standard problem for evaluating BO techniques, and the fitting of experimental current-voltage data, a classical optimization problem in solar energy research. All experiments described in this study were carried out on a DELL laptop with 64GB of RAM and an Intel Core i7-12800H processor. Unlike other state-of-the-art BO methods \cite{siemennFastBayesianOptimization2023, erikssonScalableGlobalOptimization2019, wangBatchedLargescaleBayesian2018, cowen-riversHEBOPushingLimits2022}, $SCORE$ seamlessly operates on conventional computing systems without the need for specialized supercomputing infrastructure. At a later stage, $SCORE$ shall be evaluated on additional benchmark problems against these cutting-edge techniques, which often require extensive runtimes (hours if not days) on powerful supercomputers.

\subsection{Ackley Function}
The Ackley function, which is characterized by a needle-in-a-haystack landscape with multiple local minima and one global minimum of 0, is first considered in 10D in the domain [-5, 10]. 

\begin{figure}[h]
  \centering
  \includegraphics[scale=0.5]{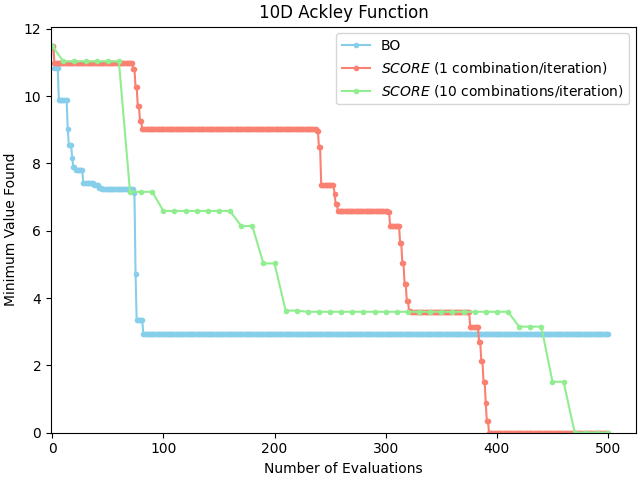}
  \caption{Convergence plots of Bayesian Optimization and two $SCORE$ configurations on the 10D Ackley function.}
  \label{fig:fig1}
\end{figure}

Starting with 20 random initial points (twice the number of dimensions), BO becomes trapped in a local minimum after roughly 100 iterations, whereas $SCORE$ successfully finds the global minimum in two different configurations, as shown in Figure \ref{fig:fig1}. 

\begin{figure}[h]
  \centering
  \includegraphics[scale=0.5]{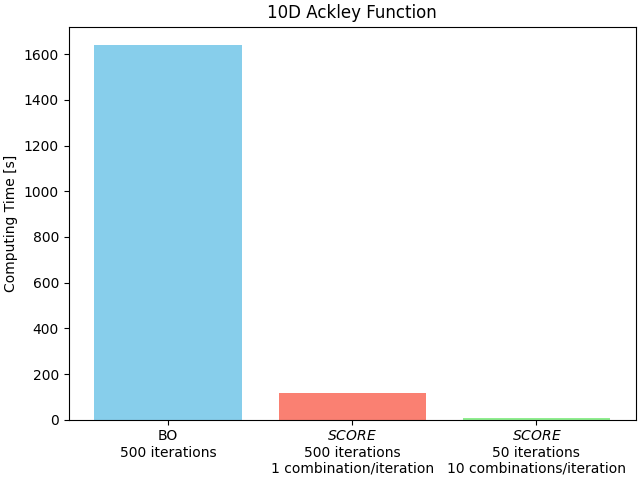}
  \caption{Average computing times of Bayesian Optimization and two $SCORE$ configurations on the 10D Ackley function.}
  \label{fig:fig2}
\end{figure}

It's important to acknowledge that the initialization process and choice of hyperparameters significantly influence the outcomes of both approaches. However, the primary focus of the results presented here is to demonstrate that, using a simple configuration, $SCORE$ can efficiently identify the global minimum at a fraction of BO’s computational cost. Moreover, for the same total number of function evaluations, $SCORE$ can perform either one evaluation per iteration or multiple evaluations simultaneously. Even though testing 10 combinations per iteration does not reduce the final number of function evaluations needed to find the global minimum (in this case), it reduces the number of times the GP surrogate model is called by a factor of 10, which further decreases the computing time, as depicted in Figure \ref{fig:fig2}.

\begin{figure}[h]
  \centering
  \includegraphics[scale=0.5]{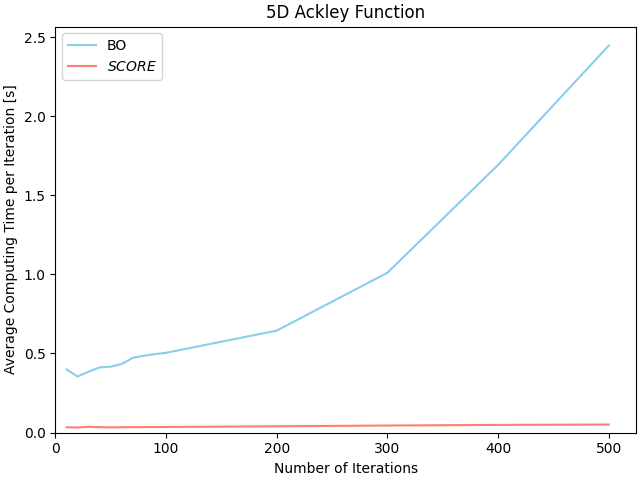}
  \caption{Computing times of Bayesian Optimization and $SCORE$ as a function of the number of iterations on the 5D Ackley function.}
  \label{fig:fig3}
\end{figure}

Most notably, $SCORE$ lifts BO's curse of dimensionality and keeps the computing time in check, even with increasing iterations, as shown in Figure \ref{fig:fig3}.

\begin{figure}[h]
  \centering
  \includegraphics[scale=0.5]{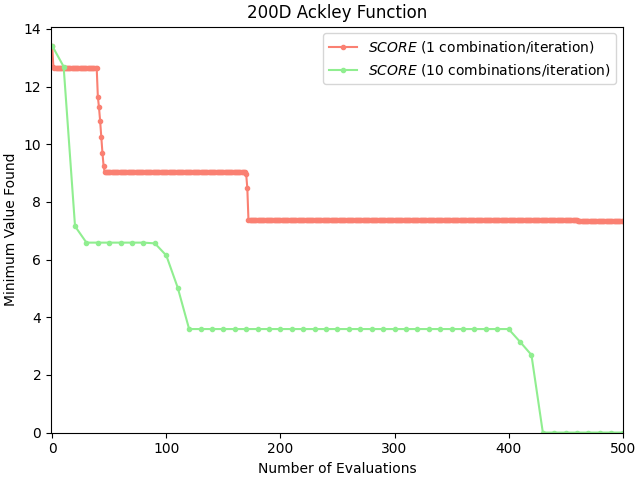}
  \caption{Convergence plots of two $SCORE$ configurations on the 200D Ackley function.}
  \label{fig:fig4}
\end{figure}

This is particularly beneficial for high-dimensional problems, which often require a higher number of initial points and iterations to find the optimal solution. Aside from limiting the number of GP regression calls and reducing computing time, the ability to choose the number of parameter combinations to assess per iteration can also expedite $SCORE$’s discovery of the global optimum, as illustrated in Figure \ref{fig:fig4} with the 200D Ackley function.

In a matter of minutes and with fewer than 500 objective function evaluations, $SCORE$ identifies the global minimum of the 200D Ackley function, a task that state-of-the-art BO techniques struggle to achieve even after running thousands of iterations for hours on powerful supercomputers \cite{erikssonScalableGlobalOptimization2019}. Even though the computing time of $SCORE$ is proportional to dimensionality, this is largely offset by $SCORE$'s ability to reduce both the time needed to perform GP regression and the number of times this process is called (by evaluating multiple parameter combinations per iteration).

\subsection{Experimental Data}
In solar energy research, established physical models exist to mimic solar cell behavior, but fitting their intricately correlated parameters can be challenging when limited information is available. A relatively simple example is the single-diode model (SDM), which is commonly used to simulate the current-voltage (IV) curve of solar cells and panels, a crucial indicator of photovoltaic (PV) performance. The SDM is described by the following equation:
\begin{equation}
I = I_L - I_o[exp( \frac{V+IR_s}{a})-1]- \frac{V + IR_s}{R_{sh}}
\end{equation}	
In this equation, five parameters are used to establish the relationship between a solar panel's output current $I$ and its operating voltage $V$ under defined operating conditions: the light-induced current $I_L$, saturation current $I_o$, series resistance $R_s$, shunt resistance $R_{sh}$, and modified ideality factor $a$.  

\begin{figure}[h]
  \centering
  \includegraphics[scale=0.5]{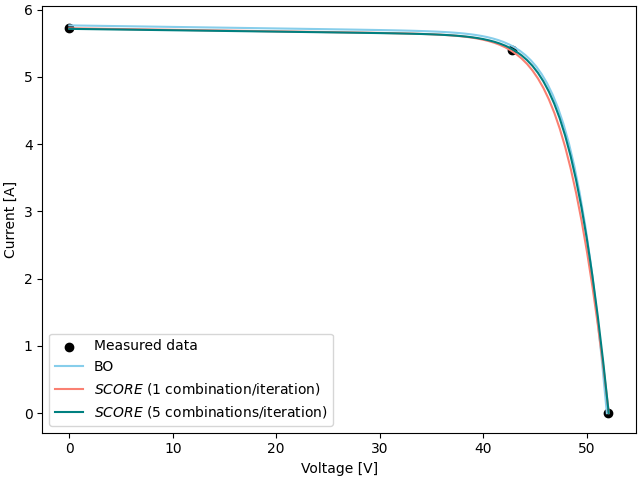}
  \caption{Measured data vs. fits obtained by Bayesian Optimization and two $SCORE$ configurations.}
  \label{fig:fig5}
\end{figure}

In real-life scenarios, only a few IV curve points are available – often the short-circuit current, maximum power point, and open-circuit voltage (the black dots in Figure \ref{fig:fig5}, from left to right, respectively). Fitting the SDM thus becomes a challenging multi-solution problem, especially due to the strong correlations between the parameters, which span different orders of magnitude. Herein, we demonstrate $SCORE$’s ability to fit the SDM to these three data points, which are taken from the datasheet of a commercial solar panel. 

Starting with one random initial point, $SCORE$ can successfully fit the measured data in under 10 function evaluations and at a fraction of the time needed by BO. However, despite the seemingly adequate fits of the IV properties, it is essential to recognize that an infinite number of solutions exist to this problem. This complexity arises because above or below certain thresholds, some SDM parameters (e.g. $R_{sh}$) start having little to no impact on the shape of the IV curve. The multi-solution nature of such real-world optimization problems highlights the importance of thoroughly navigating the parameter space and displaying more than just the optimal fit with the lowest residual error. Even though $SCORE$ optimizes in 1D, the chosen GP or probabilistic surrogate model can be called to approximate the objective function and construct probability distributions over the possible outcomes, based on the evaluated parameter combinations.

\section{Perspectives}
The technique outlined in this paper shows promising results through its ability to overcome BO’s curse of dimensionality and drastically reduce the computing time of complex optimization problems. Nonetheless, more tests are needed to fully understand $SCORE$'s robustness and evaluate its performance against state-of-the-art techniques on different benchmark problems. As for classical BO, $SCORE$ is sensitive to the initialization process as well as the choice of surrogate model, acquisition function, and hyperparameters, indicating the need for further testing. 

In addition, tailoring the surrogate model and acquisition function to each parameter may offer $SCORE$ heightened flexibility and expedite convergence toward the optimal solution. Parallelizing this sequential process could also further speed up $SCORE$’s execution time. Alternatively, when proposing the next set of parameter combinations for objective function evaluation, it may be worth computing the score of all feasible parameter combinations (based on the scores of the individual parameter values) to identify the most promising candidates. While not viable in continuous and high-dimensional spaces, this strategy can prove successful in low-dimensional discrete spaces where achieving optimal solutions does not necessitate a highly precise parameter mesh, or if it is applied iteratively to gradually refine the search space boundaries.

\section{Conclusion}
In short, a 1D reparametrization technique named $SCORE$ is proposed to break BO's curse of dimensionality. By projecting the objective function evaluations onto each parameter – akin to obtaining marginal distributions from the joint distribution – $SCORE$ partitions the BO process and executes the surrogate GP model in 1D spaces instead of the full search space. It assigns a score to the most likely values of each parameter, from which the most likely parameter combinations are prioritized for subsequent testing. This drastically reduces BO’s computational complexity and allows multiple parameter combinations to be evaluated at the same time, which further reduces the total computing time and possibly speeds up convergence. This working paper showcases $SCORE$'s ability to tackle both a complex needle-in-a-haystack optimization problem and a real-world challenge, but additional efforts are needed to assess its robustness and generalizability.\\

A Python implementation of $SCORE$ is currently under development and available at: \url{https://github.com/hi-paris/SCORE}.

\section*{Acknowledgments}
I would like to express my gratitude to my supervisors Yvan Bonnassieux and Jean-Baptiste Puel for their support. I initially designed $SCORE$ to solve solar energy research problems after realizing that BO was taking more time to run than my simulation models (i.e. my objective function evaluation), thus defeating its purpose. I expanded the scope of this approach to encompass more generic problems and drafted this working paper to demonstrate $SCORE$'s capabilities before the fast-approaching end of my PhD journey. I would also like to thank Pierre-Antoine, Machine Learning Engineer at Hi! PARIS, for developing the Python package for $SCORE$, which I hope will undergo thorough testing and benchmarking to enhance its public utility. This research work is also supported by Hi! PARIS, the Center on Data Analytics and Artificial Intelligence for Science, Business and Society created by Institut Polytechnique de Paris (IP Paris) and HEC Paris and joined by Inria (Centre Inria de Saclay).

\bibliographystyle{unsrt}  
\bibliography{references}

\end{document}